\pdfoutput=1
\documentclass[11pt]{article}

\usepackage[final]{acl}

\usepackage{times}
\usepackage{latexsym}
\usepackage{hyperref}

\usepackage[T1]{fontenc}
\usepackage[utf8]{inputenc}

\usepackage{booktabs}
\usepackage{graphicx}
\usepackage{makecell}

\usepackage{microtype}

\usepackage{inconsolata}

\title{Augmenting Lateral Thinking in Language Models with Humor and Riddle Data for the BRAINTEASER Task}

\author{Mina Ghashami\thanks{Both authors contributed equally to this work.} \\
  Amazon Web Services \\
  \texttt{ghashami@amazon.com} \\\And
  Soumya Smruti Mishra\footnotemark[1] \\
  Amazon Web Services \\
  \texttt{soumish@amazon.com} \\}

\begin{document}
\maketitle
\begin{abstract}
The SemEval 2024 BRAINTEASER task challenges language models to perform lateral thinking---a form of creative, non-linear reasoning that remains underexplored in NLP. The task comprises two subtasks, Sentence Puzzle and Word Puzzle, requiring models to defy conventional commonsense associations.

We present a system that fine-tunes DeBERTaV3 using HuggingFace's AutoModelForMultipleChoice architecture. We augment the provided training data with two additional sources: (1) a humor-style question-answering dataset generated via GPT-4 prompting, and (2) the RiddleSense dataset. This data augmentation strategy is motivated by the observation that humor and riddles share the lateral reasoning structure required by the task. Our best system achieves 92.5\% overall accuracy on the Sentence Puzzle subtask and 80.2\% on the Word Puzzle subtask, ranking 6th out of 31 teams and 10th out of 23 teams, respectively. We further show that the choice of task formulation matters: framing the problem as multiple-choice rather than sequence classification yields a 10-point accuracy improvement with the same base model. Our analysis reveals that data augmentation with humor and riddle data is particularly effective for sentence-level lateral reasoning, while word-level puzzles remain a harder challenge.
\end{abstract}

\section{Introduction}

The success of language models has inspired the Natural Language Processing community to attend to tasks that require implicit and complex reasoning. Human reasoning encompasses two broad categories: lateral and vertical thinking. Lateral thinking demands unconventional, creative reasoning that deviates from traditional logical processes and has received relatively little attention from the NLP community. Vertical thinking, on the other hand, relies on logical, step-by-step reasoning and has been the focus of considerable research in recent years.

\citet{jiang-etal-2023-brainteaser} introduced the BRAINTEASER dataset as a benchmark for evaluating lateral thinking in question-answering systems. The SemEval 2024 BRAINTEASER task \citep{jiang-ilievski-ma:2024:SemEval2024} derives a focused challenge from this dataset, comprising two subtasks: Sentence Puzzle, which requires reinterpreting sentence-level meaning, and Word Puzzle, which requires reasoning about orthographic or phonetic properties of words. In both cases, models must go beyond conventional commonsense associations and recognize non-obvious interpretations---a capability that most existing NLP benchmarks, which focus on vertical (logical, step-by-step) reasoning, do not test.

Our system uses pre-trained models---BERT \citep{devlin2019bert} and DeBERTaV3 \citep{he2023debertav3}---through HuggingFace's \citep{wolf2020huggingfaces} AutoModelForMultipleChoice and AutoModelForSequenceClassification architectures. We mix Sentence and Word Puzzle training datasets to broaden exposure to different types of lateral thinking challenges. Additionally, we augment the training data with a humor/jokes dataset generated by GPT-4 \citep{openai2024gpt4} and the RiddleSense \citep{lin-etal-2021-riddlesense} dataset, both of which require creative, non-literal reasoning similar to the target task.

Our experiments reveal several findings. First, the AutoModelForMultipleChoice formulation substantially outperforms AutoModelForSequenceClassification, confirming that task-appropriate architecture design matters even with strong pre-trained representations. Second, augmenting training data with humor and riddle datasets consistently improves performance on the Sentence Puzzle. Third, the same augmentation strategy does not transfer to the Word Puzzle, suggesting that sentence-level and word-level lateral reasoning require different data characteristics. Our system ranked 6th out of 31 teams on the Sentence Puzzle and 10th out of 23 teams on the Word Puzzle.

Our main contributions are as follows:
\begin{enumerate}
    \item We show that the multiple-choice architecture (AutoModelForMultipleChoice) significantly outperforms the sequence classification formulation for this task, yielding a 10-point accuracy gain with the same BERT base model.
    \item We demonstrate that augmenting training data with GPT-4-generated humor QA data and the RiddleSense dataset improves lateral thinking performance, achieving 92.5\% accuracy on the Sentence Puzzle.
    \item We provide an analysis of failure modes and the differential effect of data augmentation on sentence-level versus word-level lateral reasoning.
\end{enumerate}

Our code and data will be available at \url{https://github.com/soumyasmruti/semeval-2024-brainteaser} after cleaning and de-anonymization.

\section{Background}
The task involves two types of brain teasers: Sentence Puzzle and Word Puzzle. In Sentence Puzzle, the input is a sentence-based question that defies commonsense, with multiple-choice answers. For instance, "A man shaves everyday, yet keeps his beard long." The choices include "He is a barber," "He wants to maintain his appearance," and so on. The Word Puzzle involves a word-based teaser, like "What part of London is in France?" with choices focusing on letters in the words (e.g., "The letter N"). The output in both cases is the selection of the correct choice that represents lateral thinking.

In order to counter the potential for Large Language Models (LLMs) memorizing solutions, BRAINTEASER \citep{jiang-etal-2023-brainteaser} incorporates two novel methods of puzzle generation: semantic and context reconstruction. These techniques generate variations of puzzles that preserve the core challenge of overturning conventional commonsense reasoning without altering the fundamental nature of the puzzles. This approach is aimed at enhancing the robustness of the puzzles against the memorization capabilities of LLMs, ensuring that the puzzles continue to effectively test the models' ability to engage in lateral thinking by challenging ingrained commonsense assumptions. This is to ensure the model is evaluating reasoning ability rather than memorization.

Systems are evaluated based on two accuracy metrics: Instance-based Accuracy, considering each question (original and adversarial) as a separate instance, and Group-based Accuracy, where a system must correctly solve all questions in a group (original and its adversarial versions) to score.

\section{Related Work}
We can broadly categorize the reasoning landscape of language models into two groups. The first, is `commonsense reasoning`, also known as `vertical reasoning`. This refers to the ability to make deductions based on everyday knowledge. 
The second category is `lateral reasoning`; i.e. a creative problem-solving approach that involves looking at situations from unconventional perspectives. 

Researchers have explored various approaches to endow LLMs with commonsense reasoning abilities \citep{rae2021scaling}. One prominent approach is the use of knowledge graphs, which represent structured knowledge in the form of entities and their relationships \citep{ilievski2021cskg}. \citet{wang2021inductive} proposed a method for incorporating commonsense knowledge from ConceptNet \citep{speer2018conceptnet} into language models, leading to improved performance on commonsense reasoning tasks.

Another approach involves fine-tuning pre-trained LLMs on commonsense reasoning datasets. \citet{huang2019cosmos} introduced the COSMOS QA dataset, which consists of multiple-choice questions that require commonsense reasoning. They showed that fine-tuning pre-trained LLMs on this dataset can significantly improve their commonsense reasoning capabilities.

Researchers have also investigated the use of prompting techniques to elicit commonsense reasoning from LLMs without explicit fine-tuning. \citet{zhou2022conditional} proposed a method called ``Conditional Prompt-Tuning'' that enables LLMs to perform commonsense reasoning by conditioning on carefully designed prompts. In another work, \citet{wei2022chain} showed that chain-of-thought prompting can unlock LLM reasoning ability via effective prompting techniques. 

There has not been extensive research on lateral thinking in LLMs. Recently, \citet{xie2023olagpt} proposed OlaGPT, a cognitive architecture framework that summarizes various methods of human reasoning into Chain-of-Thought (CoT) templates to improve LLM reasoning.

\paragraph{Data Augmentation for Reasoning.} Data augmentation has been widely used to improve model robustness in NLP tasks. In the context of reasoning, augmenting training data with related but distinct datasets has shown promise for improving generalization. Our work extends this idea by using humor and riddle datasets as augmentation sources, motivated by the structural similarity between jokes, riddles, and lateral thinking puzzles---all require reinterpreting familiar concepts in unexpected ways.

Overall, while LLMs have shown flashes of non-linear, exploratory thinking on some benchmarks, lateral thinking as a cognitive process remains an open challenge.

\section{Methodology}
In this section, we describe different methods and approaches we employed in solving the BrainTeaser puzzle.

%    \subsection{Zero shot with GPT 4}
%    In an innovative zero-shot approach, we presented GPT-4 with /questions and answer choices extracted directly from the test dataset, without resorting to any training data or model fine-tuning. This methodology allowed us to tap into GPT-4's intrinsic understanding of language and context. By circumventing the need for explicit training or model adaptation, we leveraged GPT-4's pre-existing knowledge and capabilities to navigate through the question-answer pairs in the test data and accurately select the correct answers. 
  %/

\subsection{Sequence Classification with BERT}
In this approach, we enhanced the performance of a sequence classification model through the instruction fine-tuning process. We leveraged the powerful contextual embeddings provided by BERT \citep{devlin2019bert}. Our methodology involved initializing the model with pre-trained BERT weights and employing the streamlined `AutoModelForSequenceClassification` class from the Hugging Face Transformers library, which linearly projects the embedding from
the language model encoder to each document
into the class logits for that document. We instructed the model with selecting the most appropriate answer from a set of four choices provided alongside a given question. Despite the meticulous fine-tuning process our experimental results revealed sub-optimal performance.

\subsection{MultipleChoice QA with BERT and DeBERTa}
We leveraged the versatile `AutoModelForMultipleChoice` architecture from Hugging Face's library, which integrates a pre-trained transformer model with a specialized classification head. This architecture was pivotal in adapting the model for our multiple-choice task, which involved combining both Word Puzzle and Sentence Puzzle datasets to diversify our training data.

To ensure optimal performance, we split our training data into separate training and validation sets. Throughout the training process, we utilized the validation set to fine-tune hyperparameters, ensuring the model's efficacy.

The AutoModelForMultipleChoice architecture comprises a pre-trained base transformer augmented with a classification head. This head, typically consisting of neural network components such as linear layers and activation functions, enables the model to make informed multiple-choice predictions.

Our model initialization involved embedding pre-trained DeBERTa representations, followed by further training on the designated training dataset. This approach facilitated the model's adaptation to our specific task requirements, ultimately enhancing its performance.

\subsubsection{Augmenting with RiddleSense and Humor Data}
Next, we decided to use two additional data sources to augment our training data. This was with the aim of expanding the diversity of our dataset, enriching it with a wide range of humor styles, scenarios, and perspectives. This augmentation not only increases the robustness and variety of our model but also enhances its adaptability to different contexts. We utilized the public Riddlesense dataset as well as creating humor style data by prompting GPT 4. 

The RiddleSense dataset consists of riddles---questions presented in a cryptic or metaphorical manner that challenge the reader to find a clever or unexpected answer. Riddles share a key structural property with brain teasers: they require the solver to abandon default commonsense interpretations in favor of creative alternatives.

To create the humor-style QA data, we prompted GPT-4 to generate joke-format multiple-choice questions. Jokes frequently rely on lateral thinking, as punchlines subvert the listener's expectations by reinterpreting the setup in an unexpected way. This structural parallel to the BRAINTEASER task motivates our use of humor data as an augmentation source. The details of the data generation process are provided in Appendix~\ref{sec:humordataset}.

We then used the same AutoModelForMultipleChoice architecture and trained the model on augmented training data. 

\section{Experimental setup}

\subsection{Datasets Description}

\begin{table*}[bt!]
\centering 
\begin{tabular}{lcccccccccccc} \toprule
& \multicolumn{3}{c}{Sentence Puzzle} & \multicolumn{3}{c}{Word Puzzle} \\ \cmidrule(lr){2-4}\cmidrule(lr){5-7}
{\bf Dataset} & {\bf Train} & {\bf Validation} & {\bf Test} & {\bf Train} & {\bf Validation} & {\bf Test} \\
\midrule
Provided & 405 & 102 & 120 & 316 & 80 & 96 &  \\
Humor Data GPT4 & 211 & - & - & 211 & - & - &  \\
Riddlesense & 4531 & - & - & - & - & - &  \\
\bottomrule
\end{tabular}
\caption{Dataset Statistics, `-` means the data was not used for the stage of the task. }
\label{tab:dataset}
\end{table*}

The task dataset and additional datasets used in our approaches are detailed in Table ~\ref{tab:dataset}, with all datasets being in the English language. We did not perform any extra pre-processing on the original training or test data. To generate humor data, we used GPT-4 \citep{openai2024gpt4} using prompt engineering. Regarding the RiddleSense \citep{lin-etal-2021-riddlesense} dataset, which originally had five labels, we adapted it to a four-label format. This was achieved by reassigning questions with the fifth label as the correct answer to the fourth choice. Consequently, all fifth-choice answers across questions were remapped to their corresponding fourth choices, and all original fifth choices were discarded. Riddlesense and humor datasets, were selected for their similarity to the original training data, offering commonsense-defying puzzles. For details on the train-validation-test split, please refer to Table~\ref{tab:dataset}. We also experimented by adding SWAG \citep{zellers2018swag} and CODAH \citep{chen2019codah} datasets, but found that they reduced overall performance.

\subsection{Implementation Details}
The raw text was tokenized using a byte-level Byte-Pair Encoding (BPE) vocabulary with 50,257 merge rules, and inputs longer than 1024 tokens were truncated.

Our models were based on the BERT-base and DeBERTaV3 base architectures. The BERT model comprises 12 layers, 768-dimensional embeddings, and 12 attention heads, totaling 117M parameters. The DeBERTaV3 base model features 12 layers and a hidden size of 768, with 110M backbone parameters and a 128K token vocabulary introducing an additional 98M parameters in the embedding layer.

Both models were initialized with pre-trained weights in the AutoModelForMultipleChoice architecture. We conducted a random hyperparameter search, exploring batch sizes of [4, 16, 32] and learning rates of [5e-5, 1e-4, 2e-4]. The configurations yielding the highest validation accuracy were selected for each model size.

We utilized Amazon SageMaker for training, opting for the ml.p3.8xlarge instance for BERT-based approaches and the ml.p3.16xlarge instance for training our DeBERTaV3-based approaches. The training time for the BERT models with the original data was under 20 minutes, while the DeBERTa-based approaches were trained in under one hour. This efficient use of resources enabled us to achieve significant performance improvements with minimal cost and time. 

\section{Results}

\begin{table*}[t]
\centering \small
\begin{tabular}{lcccccccccccc} \toprule
& \multicolumn{6}{c}{Sentence Puzzle} & \multicolumn{6}{c}{Word Puzzle} \\ \cmidrule(lr){2-7}\cmidrule(lr){8-13}
{\bf Approaches} & \rotatebox{90}{\bf Original} & \rotatebox{90}{\bf Semantic} & \rotatebox{90}{\bf Context} & \rotatebox{90}{\bf Orig. + Sem.} & \rotatebox{90}{\bf Orig. + Sem. + Con.} & \rotatebox{90}{\bf Overall} & \rotatebox{90}{\bf Original} & \rotatebox{90}{\bf Semantic} & \rotatebox{90}{\bf Context} & \rotatebox{90}{\bf Orig. + Sem.} & \rotatebox{90}{\bf Orig. + Sem. + Con.} & \rotatebox{90}{\bf Overall} \\
\midrule
\textcolor{gray}{\bf Human} & \textcolor{gray}{.907} & \textcolor{gray}{.907} & \textcolor{gray}{.944} & \textcolor{gray}{.907} & \textcolor{gray}{.889} & \textcolor{gray}{.920} & \textcolor{gray}{.917} & \textcolor{gray}{.917} & \textcolor{gray}{.917} & \textcolor{gray}{.917} & \textcolor{gray}{.900} & \textcolor{gray}{.917} \\
\midrule
\textcolor{gray}{ChatGPT} & \textcolor{gray}{.608} & \textcolor{gray}{.593} & \textcolor{gray}{.679} & \textcolor{gray}{.507} & \textcolor{gray}{.397} & \textcolor{gray}{.627} & \textcolor{gray}{.561} & \textcolor{gray}{.524} & \textcolor{gray}{.518} & \textcolor{gray}{.439} & \textcolor{gray}{.292} & \textcolor{gray}{.535} \\
\midrule
\textcolor{gray}{RoBERTa-L} & \textcolor{gray}{.435} & \textcolor{gray}{.402} & \textcolor{gray}{.464} & \textcolor{gray}{.330} & \textcolor{gray}{.201} & \textcolor{gray}{.434} & \textcolor{gray}{.195} & \textcolor{gray}{.195} & \textcolor{gray}{.232} & \textcolor{gray}{.146} & \textcolor{gray}{.061} & \textcolor{gray}{.207} \\
\midrule
\makecell[l]{BERT-base +\\AMSC +\\train-data-wp+sp} & .475 & .55 & .5 & .35 & .25 & .508 & .281 & .312 & .375 & .031 & 0 & .323 \\
\midrule
\makecell[l]{BERT-base +\\AMMC +\\train-data-wp+sp} & .650 & .625 & .625 & .600 & .500 & .600 & .438 & .375 & .406 & .344 & .375 & .406 \\
\midrule
\makecell[l]{DeBERTaV3 +\\AMMC +\\train-data-wp+sp} & .900 & .900 & .850 & .900 & .825 & .883 & .75 & .75 & .625 & .719 & .500 & .708 \\
\midrule
\makecell[l]{DeBERTaV3 +\\AMMC +\\train-data-wp+sp +\\ Humor + RiddleSense} & \bf .925 & \bf .950 & \bf .900 & \bf .925 & \bf .875 & \bf .925 & - & - & - & - & - & - \\
\midrule
\makecell[l]{DeBERTaV3 +\\AMMC +\\train-data-wp +\\ Humor}& - & - & - & - & - & - & \bf .844 & \bf .812 & \bf .750 & \bf .781 & \bf .594 & \bf .802 \\
\bottomrule
\end{tabular}
\caption{SemEval2024 Task 9: BRAINTEASER results table, which shows the performance of different approaches on the test set. Orig. = Original, Sem. = Semantic, Con. = Context, AMSC = AutoModelForSequenceClassification, AMMC = AutoModelForMultipleChoice }
\label{tab:results}
\end{table*}

In Table 2, we demonstrate the performance of our model, where the provided numbers represent the accuracy for various groups. "Original," "Semantic," and "Context" denote the original question, its semantic reconstruction, and context reconstruction, respectively. These three categories are based on instance-based accuracy, where each question is treated as a separate instance. The score reports the accuracy for both the original question and its adversarial counterparts. "Orig. + Sem." represents group-based accuracy, where the original question and its semantic reconstruction are considered and calculated together. Similarly, "Orig. + Sem. + Con." includes the previous group along with the contextual reconstruction of the original question.

In the table, "AMSC" represents AutoModelForSequenceClassification, and "AMMC" represents AutoModelForMultipleChoice. The models used are bert-base-uncased and microsoft/deberta-v3-base. The notation "train-data-wp+sp" indicates that the training data for this approach includes both sentence puzzle and word puzzle training data provided by the organizers of the task. "Humor" represents the synthetic dataset generated by prompting GPT-4, and "RiddleSense" refers to the open-source RiddleSense dataset \citep{lin-etal-2021-riddlesense}. The scores of human performance and the baseline system, as provided in the original paper \citep{jiang-etal-2023-brainteaser}, are depicted in gray. Scores obtained by our system are shown in black, with the best performances for each task highlighted in bold.

\subsection{Subtask A : Sentence Puzzle}
Initially, we trained our models only on the provided sentence puzzle dataset but soon realized that combining both the sentence puzzle and word puzzle datasets yielded better validation scores. Consequently, we used the bert-base model with AutoModelForSequenceClassification, achieving an overall accuracy of 50.8\%. Given that the dataset is in a multiple-choice format, we experimented with AutoModelForMultipleChoice using the same bert model. This change significantly improved performance, increasing accuracy by 10 points to 60\%. Encouraged by this, we opted for the larger DeBERTaV3 model under the AutoModelForMultipleChoice configuration. This model, combined with the original dataset, significantly boosted performance, raising overall accuracy to 88.3\%. After incorporating additional datasets containing humor-style questions and the RiddleSense dataset, our best accuracy score reached 92.5\%. Our approach ranked 6th among the 31 teams that participated in the task and outperformed the baseline zero shot ChatGPT by almost 50 percentage points. 

\subsection{Subtask B : Word Puzzle}
The Word Puzzle setup followed a similar approach to the Sentence Puzzle. However, during the validation process, we found that the best-performing model was trained only on the original Word Puzzle training data combined with the humor dataset. Adding RiddleSense data and Sentence Puzzle data did not improve Word Puzzle validation scores, so we did not include those configurations in our final submission. This is a notable asymmetry: while Sentence Puzzle benefits from diverse augmentation, Word Puzzle does not. We hypothesize that this is because word-level puzzles rely on orthographic and phonetic manipulation rather than semantic reinterpretation, making sentence-oriented augmentation data less relevant. Our approach for this subtask ranked 10th among the 23 participating teams, but still outperformed the baseline zero-shot ChatGPT by approximately 40 percentage points.

\subsection{Ablation Study}

To isolate the contribution of each component in our system, we summarize the incremental gains on the Sentence Puzzle subtask in Table~\ref{tab:ablation}. Each row adds one component relative to the row above it.

The largest single gain (+28.3 points) comes from switching from BERT to DeBERTaV3, confirming the importance of the pre-trained model's capacity. The architectural choice between sequence classification and multiple choice also yields a substantial improvement (+9.2 points), underscoring that task formulation is a meaningful design decision, not merely an implementation detail. Data augmentation provides a further 4.2-point gain on Sentence Puzzle. For Word Puzzle, the best configuration (DeBERTaV3 + AMMC + wp + Humor) achieves 80.2\% overall accuracy, while adding RiddleSense or Sentence Puzzle data reduces performance.

\begin{table}[h]
\centering \small
\begin{tabular}{p{4.2cm}cc} \toprule
{\bf Configuration} & {\bf Acc.} & {\bf $\Delta$} \\
\midrule
BERT + AMSC + wp + sp & 50.8 & --- \\
BERT + AMMC  wp + sp & 60.0 & +9.2 \\
DeBERTaV3 + AMMC + wp + sp & 88.3 & +28.3 \\
\quad + Humor + RS & \bf 92.5 & +4.2 \\
\bottomrule
\end{tabular}
\caption{Ablation on Sentence Puzzle (overall accuracy, \%). RS = RiddleSense, AMSC = AutoModelForSequenceClassification, AMMC = AutoModelForMultipleChoice.}
\label{tab:ablation}
\end{table}

\subsection{Error Analysis}

We analyze the errors made by our best system to identify systematic failure patterns.

\paragraph{Instance-based vs.\ Group-based Accuracy:} A notable pattern across all configurations is the significant drop from instance-based to group-based accuracy. For example, on the Sentence Puzzle, our best system achieves 92.5\% on original questions but only 87.5\% on the Orig.+Sem.+Con.\ group metric. This indicates that while the model often answers the original question correctly, it fails on at least one adversarial variant, suggesting that its reasoning is not fully robust---it may be relying on surface patterns that break under semantic or contextual reconstruction.

\paragraph{Sentence Puzzle vs.\ Word Puzzle Gap:} Our system shows a 12.3-point gap between Sentence Puzzle (92.5\%) and Word Puzzle (80.2\%) overall accuracy. Sentence Puzzles require reinterpreting the semantic content of a sentence, which aligns well with the semantic representations learned by transformer models. Word Puzzles, in contrast, often require reasoning about orthographic structure (e.g., identifying that ``N'' is the part of ``London'' that is in ``France''), which is not directly captured by subword tokenization. This suggests that improving Word Puzzle performance may require architectural changes or character-level augmentation strategies rather than additional semantic training data.

\paragraph{Effect of Augmentation on Word Puzzles:} Adding RiddleSense data, which helped the Sentence Puzzle, actually decreased Word Puzzle validation scores. RiddleSense riddles are primarily semantic in nature, which may introduce noise when the target task requires character-level or phonetic reasoning. This finding highlights that data augmentation must be carefully matched to the reasoning type required by the target task.

\section{Conclusion}

We presented a system for the SemEval 2024 BRAINTEASER task that achieves 92.5\% accuracy on the Sentence Puzzle (6th/31 teams) and 80.2\% on the Word Puzzle (10th/23 teams). Our results demonstrate three key findings: (1) the multiple-choice architecture is substantially better suited to this task than sequence classification, (2) augmenting training data with humor and riddle datasets effectively improves sentence-level lateral reasoning, and (3) data augmentation benefits do not transfer uniformly across reasoning types---word-level puzzles require different strategies. The gap between instance-based and group-based accuracy across all systems highlights that adversarial robustness remains a challenge for lateral thinking tasks. Future work should explore character-level representations or tokenization-aware approaches for word puzzles, larger-scale synthetic data generation with better quality control, and ensemble methods that combine semantic and orthographic reasoning capabilities.

\section{Limitations}

Our work has several limitations that should be considered when interpreting the results.

\paragraph{Small Training Data:} The provided training data is small (405 Sentence Puzzle and 316 Word Puzzle instances). While our augmentation strategy mitigates this to some extent, the limited data constrains the generalizability of our conclusions about which augmentation strategies are most effective.

\paragraph{Base-size Models Only:} We experimented exclusively with base-size models (BERT-base, DeBERTaV3-base) due to computational constraints. Larger models or recent LLM-based approaches (e.g., instruction-tuned models with in-context learning) may exhibit different augmentation dynamics and could potentially close the gap on Word Puzzles.

\paragraph{Synthetic Data Quality:} Our humor dataset was generated by prompting GPT-4 with relatively simple prompts and manually checked for duplicates. A more systematic generation and filtering pipeline could yield higher-quality and more diverse augmentation data. Additionally, the humor data generation process is not fully reproducible, as GPT-4 outputs are non-deterministic.

\paragraph{Limited Analysis of Adversarial Robustness:} While we report group-based accuracy metrics that capture robustness to semantic and context reconstruction, we do not perform a detailed analysis of which types of adversarial perturbations are most challenging. A finer-grained analysis could inform more targeted approaches to improving robustness.

\paragraph{Single Task Evaluation:} Our approach is evaluated solely on the BRAINTEASER task. It remains an open question whether the data augmentation strategies we propose generalize to other lateral thinking or creative reasoning benchmarks.

% \section*{Acknowledgements}

% Bibliography entries for the entire Anthology, followed by custom entries
%\bibliography{anthology,custom}
% Custom bibliography entries only

% \bibliographystyle{plain}
\bibliography{latex/acl_latex}

@inproceedings{jiang-etal-2023-brainteaser,
    title = "{BRAINTEASER}: Lateral Thinking Puzzles for Large Language Models",
    author = "Jiang, Yifan  and
      Ilievski, Filip  and
      Ma, Kaixin  and
      Sourati, Zhivar",
    editor = "Bouamor, Houda  and
      Pino, Juan  and
      Bali, Kalika",
    booktitle = "Proceedings of the 2023 Conference on Empirical Methods in Natural Language Processing",
    month = dec,
    year = "2023",
    address = "Singapore",
    publisher = "Association for Computational Linguistics",
    url = "https://aclanthology.org/2023.emnlp-main.885",
    doi = "10.18653/v1/2023.emnlp-main.885",
    pages = "14317--14332",
}

@misc{wolf2020huggingfaces,
      title={HuggingFace's Transformers: State-of-the-art Natural Language Processing}, 
      author={Thomas Wolf and Lysandre Debut and Victor Sanh and Julien Chaumond and Clement Delangue and Anthony Moi and Pierric Cistac and Tim Rault and Rémi Louf and Morgan Funtowicz and Joe Davison and Sam Shleifer and Patrick von Platen and Clara Ma and Yacine Jernite and Julien Plu and Canwen Xu and Teven Le Scao and Sylvain Gugger and Mariama Drame and Quentin Lhoest and Alexander M. Rush},
      year={2020},
      eprint={1910.03771},
      archivePrefix={arXiv},
      primaryClass={cs.CL}
}

@inproceedings{lin-etal-2021-riddlesense,
    title = "RiddleSense: Reasoning about Riddle Questions Featuring Linguistic Creativity and Commonsense Knowledge",
    author = "Lin, Bill Yuchen and Wu, Ziyi and Yang, Yichi and Lee, Dong-Ho and Ren, Xiang",
    booktitle = "Proceedings of the 59th Annual Meeting of the Association for Computational Linguistics (ACL-IJCNLP 2021): Findings",
    year = "2021",
    note={to appear}
}

@article{wei2022chain,
  title={Chain-of-thought prompting elicits reasoning in large language models},
  author={Wei, Jason and Wang, Xuezhi and Schuurmans, Dale and Bosma, Maarten and Xia, Fei and Chi, Ed and Le, Quoc V and Zhou, Denny and others},
  journal={Advances in Neural Information Processing Systems},
  volume={35},
  pages={24824--24837},
  year={2022}
}

@article{rae2021scaling,
  title={Scaling language models: Methods, analysis \& insights from training gopher},
  author={Rae, Jack W and Borgeaud, Sebastian and Cai, Trevor and Millican, Katie and Hoffmann, Jordan and Song, Francis and Aslanides, John and Henderson, Sarah and Ring, Roman and Young, Susannah and others},
  journal={arXiv preprint arXiv:2112.11446},
  year={2021}
}

@article{xie2023olagpt,
  title={OlaGPT: Empowering LLMs With Human-like Problem-Solving Abilities},
  author={Xie, Yuanzhen and Xie, Tao and Lin, Mingxiong and Wei, WenTao and Li, Chenglin and Kong, Beibei and Chen, Lei and Zhuo, Chengxiang and Hu, Bo and Li, Zang},
  journal={arXiv preprint arXiv:2305.16334},
  year={2023}
}

@inproceedings{zhou2022conditional,
  title={Conditional prompt learning for vision-language models},
  author={Zhou, Kaiyang and Yang, Jingkang and Loy, Chen Change and Liu, Ziwei},
  booktitle={Proceedings of the IEEE/CVF Conference on Computer Vision and Pattern Recognition},
  pages={16816--16825},
  year={2022}
}

@article{huang2019cosmos,
  title={Cosmos QA: Machine reading comprehension with contextual commonsense reasoning},
  author={Huang, Lifu and Bras, Ronan Le and Bhagavatula, Chandra and Choi, Yejin},
  journal={arXiv preprint arXiv:1909.00277},
  year={2019}
}

@inproceedings{ilievski2021cskg,
  title={Cskg: The commonsense knowledge graph},
  author={Ilievski, Filip and Szekely, Pedro and Zhang, Bin},
  booktitle={The Semantic Web: 18th International Conference, ESWC 2021, Virtual Event, June 6--10, 2021, Proceedings 18},
  pages={680--696},
  year={2021},
  organization={Springer}
}

@inproceedings{wang2021inductive,
  title={Inductive learning on commonsense knowledge graph completion},
  author={Wang, Bin and Wang, Guangtao and Huang, Jing and You, Jiaxuan and Leskovec, Jure and Kuo, C-C Jay},
  booktitle={2021 International Joint Conference on Neural Networks (IJCNN)},
  pages={1--8},
  year={2021},
  organization={IEEE}
}

@article{zellers2018swag,
  title={Swag: A large-scale adversarial dataset for grounded commonsense inference},
  author={Zellers, Rowan and Bisk, Yonatan and Schwartz, Roy and Choi, Yejin},
  journal={arXiv preprint arXiv:1808.05326},
  year={2018}
}

@article{chen2019codah,
  title={Codah: An adversarially authored question-answer dataset for common sense},
  author={Chen, Michael and D'Arcy, Mike and Liu, Alisa and Fernandez, Jared and Downey, Doug},
  journal={arXiv preprint arXiv:1904.04365},
  year={2019}
}

@InProceedings{jiang-ilievski-ma:2024:SemEval2024,
  author    = {Jiang, Yifan  and  Ilievski, Filip  and  Ma, Kaixin},
  title     = {SemEval-2024 Task 9: BRAINTEASER: A Novel Task Defying Common Sense},
  booktitle      = {Proceedings of the 18th International Workshop on Semantic Evaluation (SemEval-2024)},
  month          = {June},
  year           = {2024},
  address        = {Mexico City, Mexico},
  publisher      = {Association for Computational Linguistics},
  pages     = {1996--2010},
  url       = {https://aclanthology.org/2024.semeval2024-1.271}
}

@misc{devlin2019bert,
      title={BERT: Pre-training of Deep Bidirectional Transformers for Language Understanding}, 
      author={Jacob Devlin and Ming-Wei Chang and Kenton Lee and Kristina Toutanova},
      year={2019},
      eprint={1810.04805},
      archivePrefix={arXiv},
      primaryClass={cs.CL}
}

@misc{he2023debertav3,
      title={DeBERTaV3: Improving DeBERTa using ELECTRA-Style Pre-Training with Gradient-Disentangled Embedding Sharing}, 
      author={Pengcheng He and Jianfeng Gao and Weizhu Chen},
      year={2023},
      eprint={2111.09543},
      archivePrefix={arXiv},
      primaryClass={cs.CL}
}

@misc{openai2024gpt4,
      title={GPT-4 Technical Report}, 
      author={OpenAI and Josh Achiam and Steven Adler and Sandhini Agarwal and Lama Ahmad and Ilge Akkaya and Florencia Leoni Aleman and Diogo Almeida and Janko Altenschmidt and Sam Altman and Shyamal Anadkat and Red Avila and Igor Babuschkin and Suchir Balaji and Valerie Balcom and Paul Baltescu and Haiming Bao and Mohammad Bavarian and Jeff Belgum and Irwan Bello and Jake Berdine and Gabriel Bernadett-Shapiro and Christopher Berner and Lenny Bogdonoff and Oleg Boiko and Madelaine Boyd and Anna-Luisa Brakman and Greg Brockman and Tim Brooks and Miles Brundage and Kevin Button and Trevor Cai and Rosie Campbell and Andrew Cann and Brittany Carey and Chelsea Carlson and Rory Carmichael and Brooke Chan and Che Chang and Fotis Chantzis and Derek Chen and Sully Chen and Ruby Chen and Jason Chen and Mark Chen and Ben Chess and Chester Cho and Casey Chu and Hyung Won Chung and Dave Cummings and Jeremiah Currier and Yunxing Dai and Cory Decareaux and Thomas Degry and Noah Deutsch and Damien Deville and Arka Dhar and David Dohan and Steve Dowling and Sheila Dunning and Adrien Ecoffet and Atty Eleti and Tyna Eloundou and David Farhi and Liam Fedus and Niko Felix and Simón Posada Fishman and Juston Forte and Isabella Fulford and Leo Gao and Elie Georges and Christian Gibson and Vik Goel and Tarun Gogineni and Gabriel Goh and Rapha Gontijo-Lopes and Jonathan Gordon and Morgan Grafstein and Scott Gray and Ryan Greene and Joshua Gross and Shixiang Shane Gu and Yufei Guo and Chris Hallacy and Jesse Han and Jeff Harris and Yuchen He and Mike Heaton and Johannes Heidecke and Chris Hesse and Alan Hickey and Wade Hickey and Peter Hoeschele and Brandon Houghton and Kenny Hsu and Shengli Hu and Xin Hu and Joost Huizinga and Shantanu Jain and Shawn Jain and Joanne Jang and Angela Jiang and Roger Jiang and Haozhun Jin and Denny Jin and Shino Jomoto and Billie Jonn and Heewoo Jun and Tomer Kaftan and Łukasz Kaiser and Ali Kamali and Ingmar Kanitscheider and Nitish Shirish Keskar and Tabarak Khan and Logan Kilpatrick and Jong Wook Kim and Christina Kim and Yongjik Kim and Jan Hendrik Kirchner and Jamie Kiros and Matt Knight and Daniel Kokotajlo and Łukasz Kondraciuk and Andrew Kondrich and Aris Konstantinidis and Kyle Kosic and Gretchen Krueger and Vishal Kuo and Michael Lampe and Ikai Lan and Teddy Lee and Jan Leike and Jade Leung and Daniel Levy and Chak Ming Li and Rachel Lim and Molly Lin and Stephanie Lin and Mateusz Litwin and Theresa Lopez and Ryan Lowe and Patricia Lue and Anna Makanju and Kim Malfacini and Sam Manning and Todor Markov and Yaniv Markovski and Bianca Martin and Katie Mayer and Andrew Mayne and Bob McGrew and Scott Mayer McKinney and Christine McLeavey and Paul McMillan and Jake McNeil and David Medina and Aalok Mehta and Jacob Menick and Luke Metz and Andrey Mishchenko and Pamela Mishkin and Vinnie Monaco and Evan Morikawa and Daniel Mossing and Tong Mu and Mira Murati and Oleg Murk and David Mély and Ashvin Nair and Reiichiro Nakano and Rajeev Nayak and Arvind Neelakantan and Richard Ngo and Hyeonwoo Noh and Long Ouyang and Cullen O'Keefe and Jakub Pachocki and Alex Paino and Joe Palermo and Ashley Pantuliano and Giambattista Parascandolo and Joel Parish and Emy Parparita and Alex Passos and Mikhail Pavlov and Andrew Peng and Adam Perelman and Filipe de Avila Belbute Peres and Michael Petrov and Henrique Ponde de Oliveira Pinto and Michael and Pokorny and Michelle Pokrass and Vitchyr H. Pong and Tolly Powell and Alethea Power and Boris Power and Elizabeth Proehl and Raul Puri and Alec Radford and Jack Rae and Aditya Ramesh and Cameron Raymond and Francis Real and Kendra Rimbach and Carl Ross and Bob Rotsted and Henri Roussez and Nick Ryder and Mario Saltarelli and Ted Sanders and Shibani Santurkar and Girish Sastry and Heather Schmidt and David Schnurr and John Schulman and Daniel Selsam and Kyla Sheppard and Toki Sherbakov and Jessica Shieh and Sarah Shoker and Pranav Shyam and Szymon Sidor and Eric Sigler and Maddie Simens and Jordan Sitkin and Katarina Slama and Ian Sohl and Benjamin Sokolowsky and Yang Song and Natalie Staudacher and Felipe Petroski Such and Natalie Summers and Ilya Sutskever and Jie Tang and Nikolas Tezak and Madeleine B. Thompson and Phil Tillet and Amin Tootoonchian and Elizabeth Tseng and Preston Tuggle and Nick Turley and Jerry Tworek and Juan Felipe Cerón Uribe and Andrea Vallone and Arun Vijayvergiya and Chelsea Voss and Carroll Wainwright and Justin Jay Wang and Alvin Wang and Ben Wang and Jonathan Ward and Jason Wei and CJ Weinmann and Akila Welihinda and Peter Welinder and Jiayi Weng and Lilian Weng and Matt Wiethoff and Dave Willner and Clemens Winter and Samuel Wolrich and Hannah Wong and Lauren Workman and Sherwin Wu and Jeff Wu and Michael Wu and Kai Xiao and Tao Xu and Sarah Yoo and Kevin Yu and Qiming Yuan and Wojciech Zaremba and Rowan Zellers and Chong Zhang and Marvin Zhang and Shengjia Zhao and Tianhao Zheng and Juntang Zhuang and William Zhuk and Barret Zoph},
      year={2024},
      eprint={2303.08774},
      archivePrefix={arXiv},
      primaryClass={cs.CL}
}

@misc{speer2018conceptnet,
      title={ConceptNet 5.5: An Open Multilingual Graph of General Knowledge}, 
      author={Robyn Speer and Joshua Chin and Catherine Havasi},
      year={2018},
      eprint={1612.03975},
      archivePrefix={arXiv},
      primaryClass={cs.CL}
}

\appendix

\section{Humor Dataset Details}
\label{sec:humordataset}
We used the following prompts to generate the humor-style dataset. We experimented with multiple prompts and gathered all outputs in a JSON file, then analyzed them manually.

PROMPT 1 - Could you create a dataset for me that includes humor-styled questions, each with multiple choices and an answer? The dataset should be in JSON format.

PROMPT 2 - Could you create a dataset of 40 jokes for me in JSON format? Each joke should include four options and the correct answer.

PROMPT 3 - Could you generate an additional 20 jokes with multiple choices and an answer? Please ensure there are no duplicates and that none of them are the same as those previously generated.

We initially prompted GPT-4 to generate 200 questions at once, but the output contained duplicate questions after approximately 15--16 unique ones, as the model began repeating itself. We therefore used PROMPT 3 iteratively to generate data in smaller batches, checking for duplicates manually before each addition. Below are examples of the generated jokes.

\begin{quote}
    {
        "joke": "Why did the bicycle fall over?",
        "options": [
            "A. Because it was two-tired.",
            "B. It had a flat.",
            "C. It was unbalanced.",
            "D. It slipped."
        ],
        "answer": "A"
    }
    
    {
        "joke": "What's orange and sounds like a parrot?",
        "options": [
            "A. A carrot",
            "B. An orange bird",
            "C. A tangerine",
            "D. A flamingo"
        ],
        "answer": "A"
    },
\end{quote}
\end{document}